\documentclass[final]{cvpr}

\usepackage{times}
\usepackage{epsfig}
\usepackage{graphicx}
\usepackage{amsmath}
\usepackage{amssymb}

\usepackage{pifont}
\usepackage{graphicx}
\usepackage{subfigure}
\usepackage{multirow}
\usepackage{booktabs}
\usepackage{makecell}

\usepackage[pagebackref=true,breaklinks=true,colorlinks,bookmarks=false]{hyperref}

\def\cvprPaperID{1390} 
\def\confYear{CVPR 2021}

\begin{document}

\title{Distilling Knowledge via Knowledge Review}

\author{Pengguang Chen$^{1}$ \quad Shu Liu$^{2}$ \quad Hengshuang Zhao$^{3}$ \quad Jiaya Jia$^{1,2}$  \\
	The Chinese University of Hong Kong$^{1}$\quad SmartMore$^{2}$\quad University of Oxford$^{3}$\\
	\{pgchen, leojia\}@cse.cuhk.edu.hk \quad liushuhust@gmail.com \quad hengshuang.zhao@eng.ox.ac.uk
}

\maketitle
\pagestyle{empty}
\thispagestyle{empty}

\begin{abstract}
Knowledge distillation transfers knowledge from the teacher network to the student one, with the goal of greatly improving the performance of the student network. Previous methods mostly focus on proposing feature transformation and loss functions between the same level's features to improve the effectiveness. We differently study the factor of connection path cross levels between teacher and student networks, and reveal its great importance. For the first time in knowledge distillation, cross-stage connection paths are proposed. Our new review mechanism is effective and structurally simple. Our finally designed nested and compact framework requires negligible computation overhead, and outperforms other methods on a variety of tasks. We apply our method to classification, object detection, and instance segmentation tasks. All of them witness significant student network performance improvement. Code is available at \url{https://github.com/Jia-Research-Lab/ReviewKD}
\end{abstract}

\section{Introduction}
Deep convolution neural networks (CNNs) have achieved remarkable success in a variety of computer vision tasks. However, the success of CNN is often accompanied with considerable computation and memory consumption, making it a challenging topic to apply to devices with limited resource. 
There have been techniques for training fast and compact neural networks, including designing new architectures \cite{mobilev3,fpnas,mobilenet,mobilev2}, network pruning \cite{thinet,prunning,autoprune,lottery,rethinkprune}, quantization \cite{quantization} , and knowledge distillation \cite{kd,fitnet}. 

We focus on knowledge distillation in this paper considering its practicality, efficiency, and most importantly the potential to be useful. It forms a very general line, applicable to almost all network architectures and can combine with many other strategies, such as network pruning and quantization \cite{quanmimic}, to further improve network design.

Knowledge distillation is first proposed in \cite{kd}. The process is to train a small network (also known as the student) under the supervision of a larger network (a.k.a. the teacher). In \cite{kd}, knowledge is distilled though the teacher's logit, which means the student is supervised by both ground truth labels and teacher's logits. Recently, effort has been made to improve distillation effectiveness. FitNet \cite{fitnet} distilled knowledge though intermediate features. AT \cite{at} further optimized FitNet and used the attention map of features to deliver knowledge. PKT \cite{pkt} modeled knowledge of the teacher as a probability distribution while CRD \cite{crd} used a contrastive objective to transfer knowledge. All these solutions focused on transformation and loss functions.

\begin{figure}[t]
	\centering
	\includegraphics[width=\linewidth]{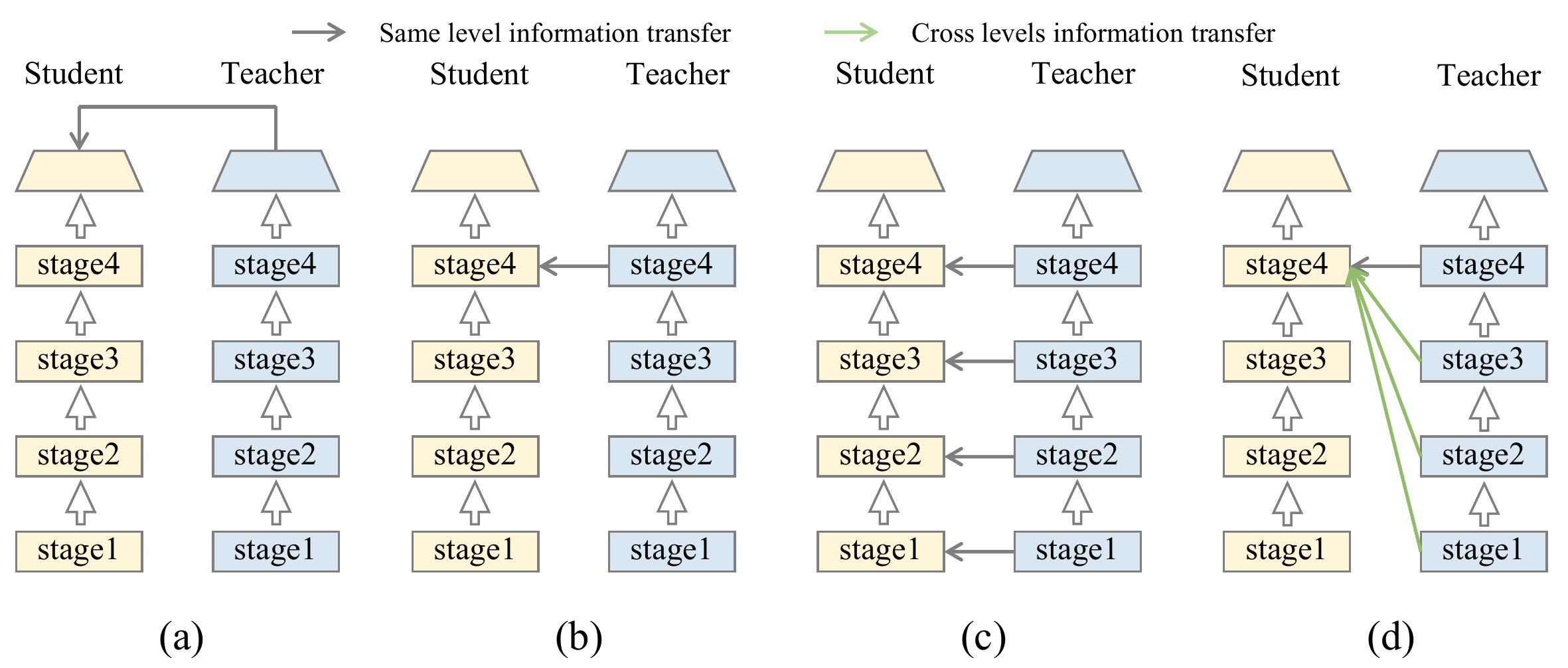}
	\caption{(a)-(c) Previous knowledge distillation frameworks. They only transfer knowledge within the same levels. (d) Our proposed ``knowledge review'' mechanism. We use multiple layers of the teacher to supervise one layer in the student. Thus, knowledge passing arises among different levels.}
	\label{fig:pipeline}
\end{figure}

\vspace{-0.1in} \paragraph{Our New Finding} We in this paper tackle this challenging problem from a new perspective regarding the connection path between the teacher and student. To briefly understand our idea, we first show how previous work deals with these paths. As shown in Figure \ref{fig:pipeline}(a)-(c), {\it all} previous methods only use the-same-level information to guide the student. For example, when supervising the student's fourth-stage output, {\it always} the teacher's fourth-stage information is utilized. This procedure looks intuitive and easy to construct. But we intriguingly reveal that it is in fact a bottleneck in the whole knowledge distillation framework -- {\it quick update of the structure surprisingly improves the whole-system performance consistently for many tasks}.

We investigate the previously neglected importance of designing connection paths in knowledge distillation and propose a new effective framework accordingly. The key modification is to use low-level features in the teacher network to supervise deeper features for the student, which results in much improved overall performance. 

We further analyze the network structure and discover the fact that the student high-level stage has the great capacity to learn useful information from the teacher's low-level features. More analysis is provided in Section \ref{sec:analysis}. This process is analogous to human learning curve \cite{learningcurve} where a young kid can only comprehend a small portion of knowledge that is taught. During the course of grow-up, more and more knowledge from past years may be gradually understood and remembered as experience.  

\vspace{-0.1in} \paragraph{Our Knowledge Review Framework} Based on these discoveries, we propose to use multi-level information of the teacher to guide one-level learning of the student network. Our novel pipeline is shown in Figure \ref{fig:pipeline}(d), which we call ``knowledge review''. 
The review mechanism is to use previous (shallower) features to guide the current feature.
It means a student has to always check what has been studied before for refreshing understanding and context of ``old knowledge''. It is a common practice for our human study to connect knowledge taught at different stages during a period of time of study.

However, how to extract useful information from multi-level information from the teacher and how to transfer them to the student are open and challenge problems. To tackle them, we propose a residual learning framework to make the learning process stable and efficient. Further, a novel attention based fusion (ABF) module and a hierarchical context loss (HCL) function are designed to boost performance. Our proposed framework makes the student network much improve the effectiveness of learning. 

By applying this idea, we achieve better performance in many computer vision tasks.
Extensive experiments in Sec. \ref{sec:experiments} manifest the vast advantage of our proposed knowledge review strategy.

\vspace{-0.1in} \paragraph{Main Contributions} 
\begin{itemize}\vspace{-0.05in} 
	\item We propose a new review mechanism in knowledge distillation, utlizing multi-level information of the teacher to guide one-level learning of the student net.\vspace{-0.1in} 
	\item We propose a residual learning framework to better realize the learning process of the review mechanism.\vspace{-0.1in} 
	\item To further improve the knowledge review mechanism, we propose an attentation based fusion (ABF) module and a hierarchical context loss (HCL) function.\vspace{-0.1in} 
	\item We achieve state-of-the-art performance of many compact models in multiple computer vision tasks by applying our distillation framework.
\end{itemize}

\section{Related Work}
Knowledge distillation concept was proposed in \cite{kd}, where the student network learns from both the ground-truth labels and the soft-labels provided by the teacher. 
FitNet \cite{fitnet} distilled knowledge through one stage intermediate feature. The idea in FitNet is simple, where the student network feature is transferred to the same shape of the teacher though convolution layers. $\mathcal{L}_2$ distance is used to measure the distance between them. 

Many methods follow FitNet and use one-stage feature to distill knowledge. 
PKT \cite{pkt} modeled knowledge of the teacher as a probability distribution and used KL divergence to measure the distance. 
RKD \cite{rkd} used multiple example relation to guide learning of the student. 
CRD\cite{crd} combined contrastive learning and knowledge distillation, and used a contrastive objective to transfer knowledge.

There are also methods using multi-stage information to transfer knowledge. 
AT \cite{at} used multiple layer attention maps to transfer knowledge. 
FSP \cite{fsp} generated FSP matrix from layer feature and used the matrix to guide the student. 
SP \cite{spkd} further improved AT. Instead of single input information, SP uses the similarity between examples to guide the student. 
OFD \cite{ofd} contained a new distance function to distill major information between the teacher and student using marginal ReLU. 

All previous methods do not discuss the possibility to ``review knowledge'', which, however, is found in our work very effective to quickly improve system performance.

\begin{figure*} [tb]
	\centering
	
	\begin{minipage}[t]{0.8\linewidth}
		\includegraphics[width=\linewidth]{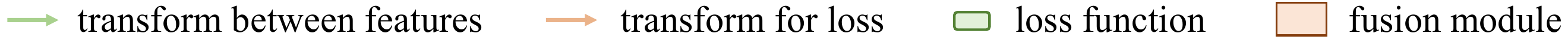}
	\end{minipage}	
	
	\subfigure[]{
		\begin{minipage}[t]{0.44\linewidth}
			\includegraphics[width=\linewidth]{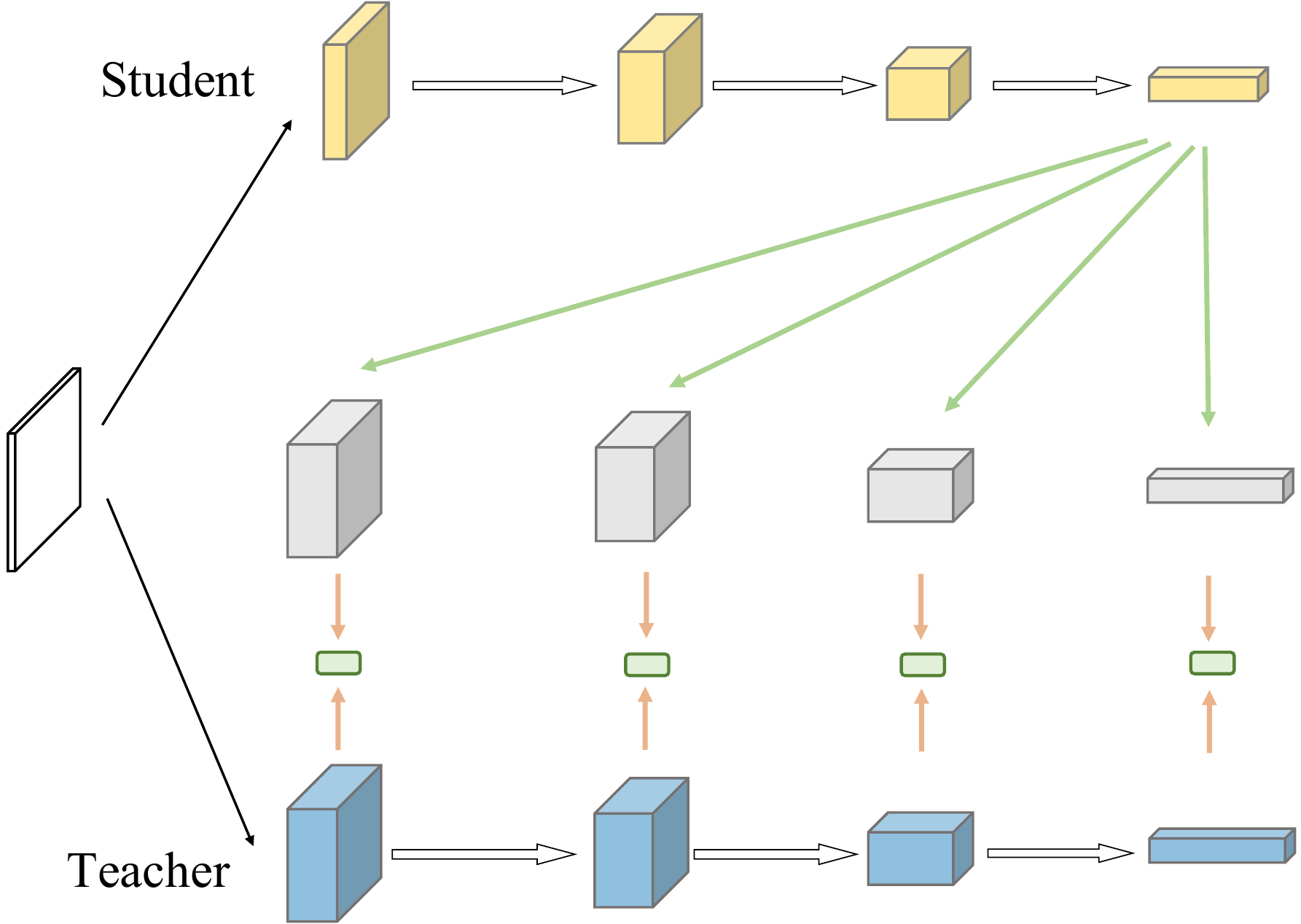}
			\label{fig:imp1}	
		\end{minipage}
	}
	\hspace{0.4in}
	\subfigure[]{
		\begin{minipage}[t]{0.44\linewidth}
			\includegraphics[width=\linewidth]{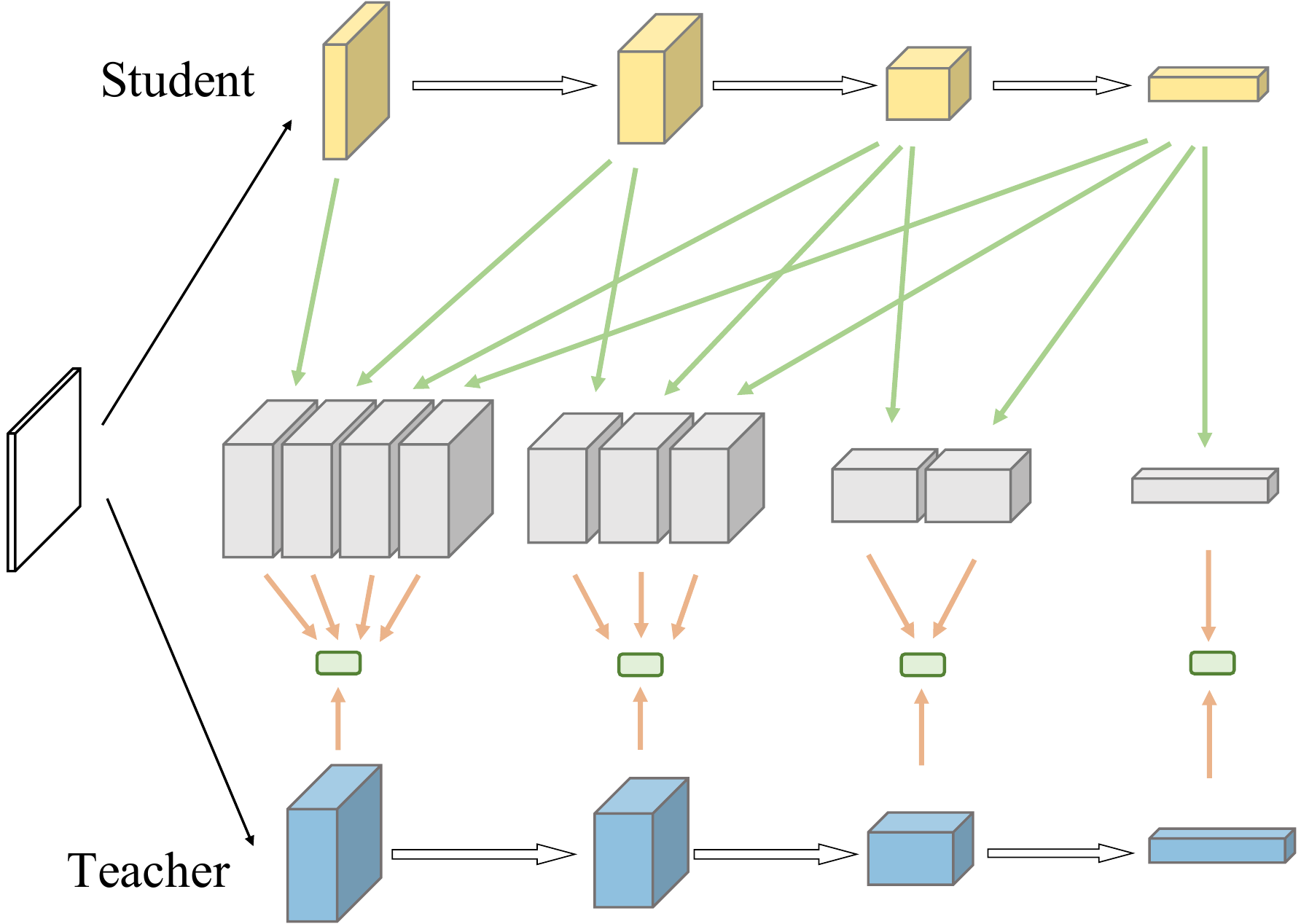}
			\label{fig:imp2}		
		\end{minipage}
	}

	\subfigure[]{
		\begin{minipage}[t]{0.44\linewidth}
			\includegraphics[width=\linewidth]{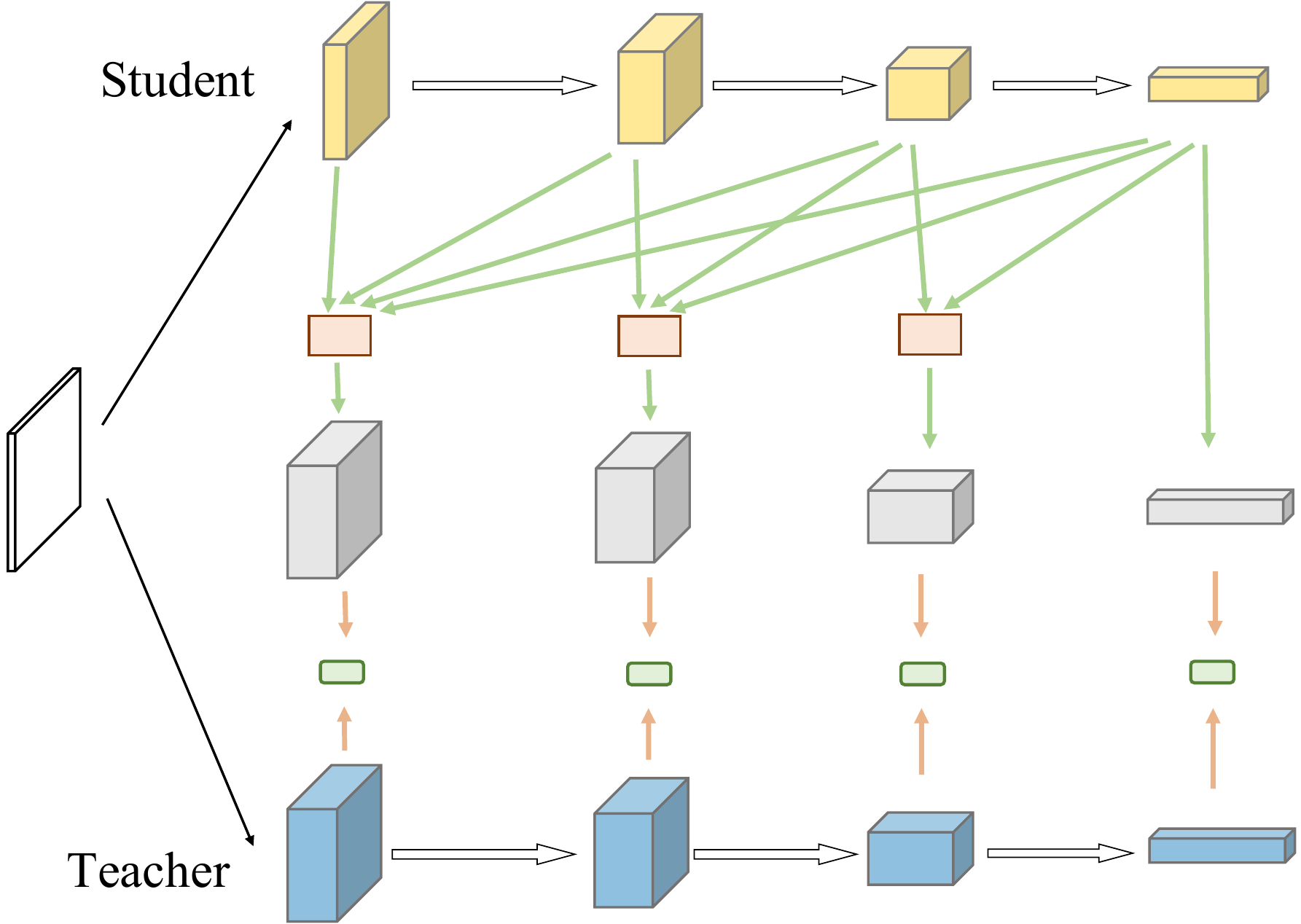}
			\label{fig:imp3}		
		\end{minipage}
	}
	\hspace{0.4in}
	\subfigure[]{
		\begin{minipage}[t]{0.44\linewidth}
			\includegraphics[width=\linewidth]{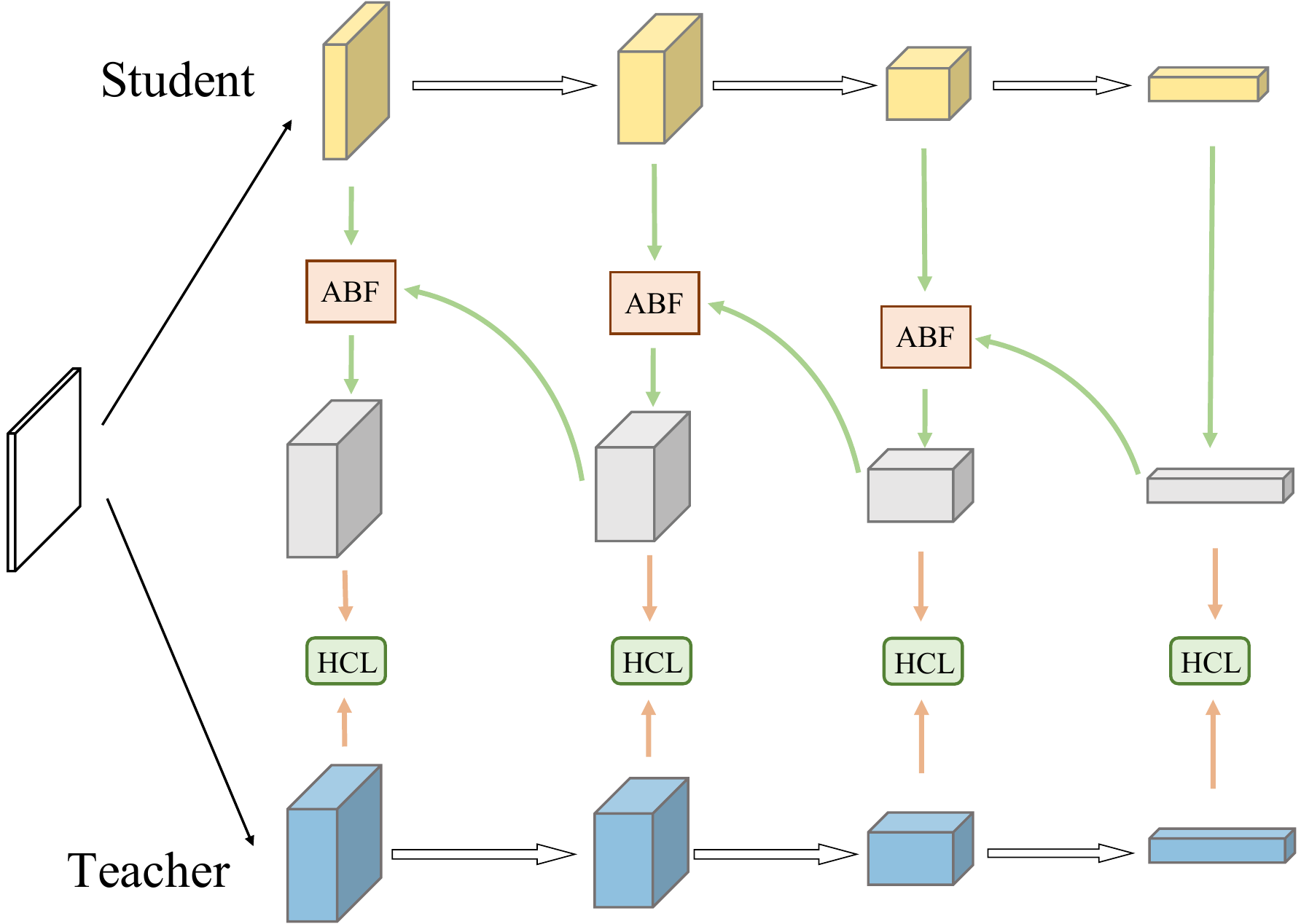}
			\label{fig:imp4}		
		\end{minipage}
	}
	
	\caption{(a) Architecture for supervising one layer of the student according to the review mechanism. (b) Direct generalization from one layer to muliple ones. The process is straightforward but costly. (c) The architecture in (b) is optimized with fusion modules to obtain a compact framework. (d) We further improved the procudure in a progressive manner and utlize redisual learning as our final architecture. Structures of ABF and HCL are in Figure \ref{fig:detail}. This figure is best viewed in color.}
	\label{fig:imp}
\end{figure*}

\begin{figure} [tb]
	\centering
	
	\subfigure[]{
		\begin{minipage}[t]{0.42\linewidth}
			\includegraphics[width=\linewidth]{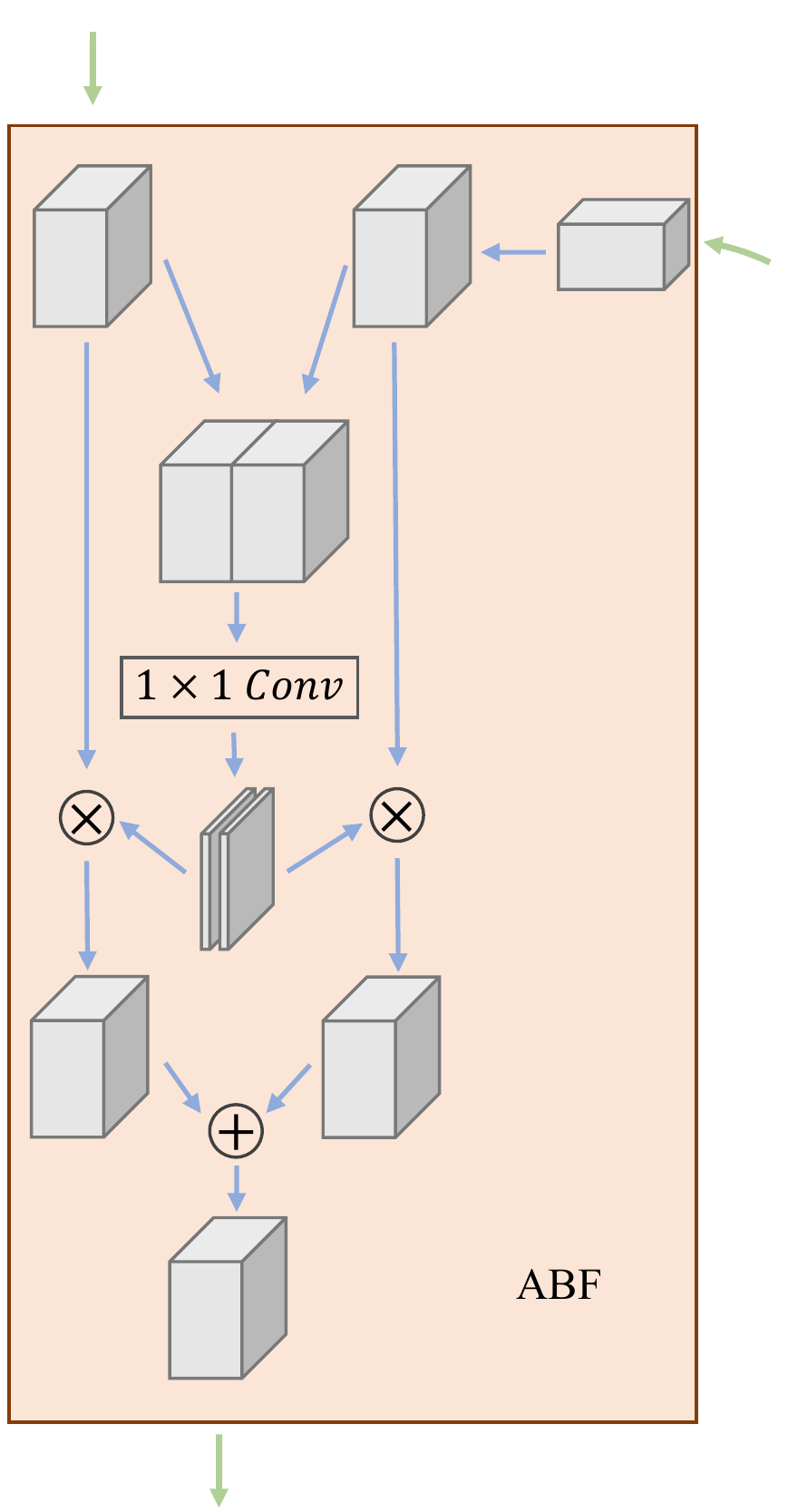}
			\label{fig:abf}	
		\end{minipage}
	}
	\hspace{0.01in}
	\subfigure[]{
		\begin{minipage}[t]{0.42\linewidth}
			\includegraphics[width=\linewidth]{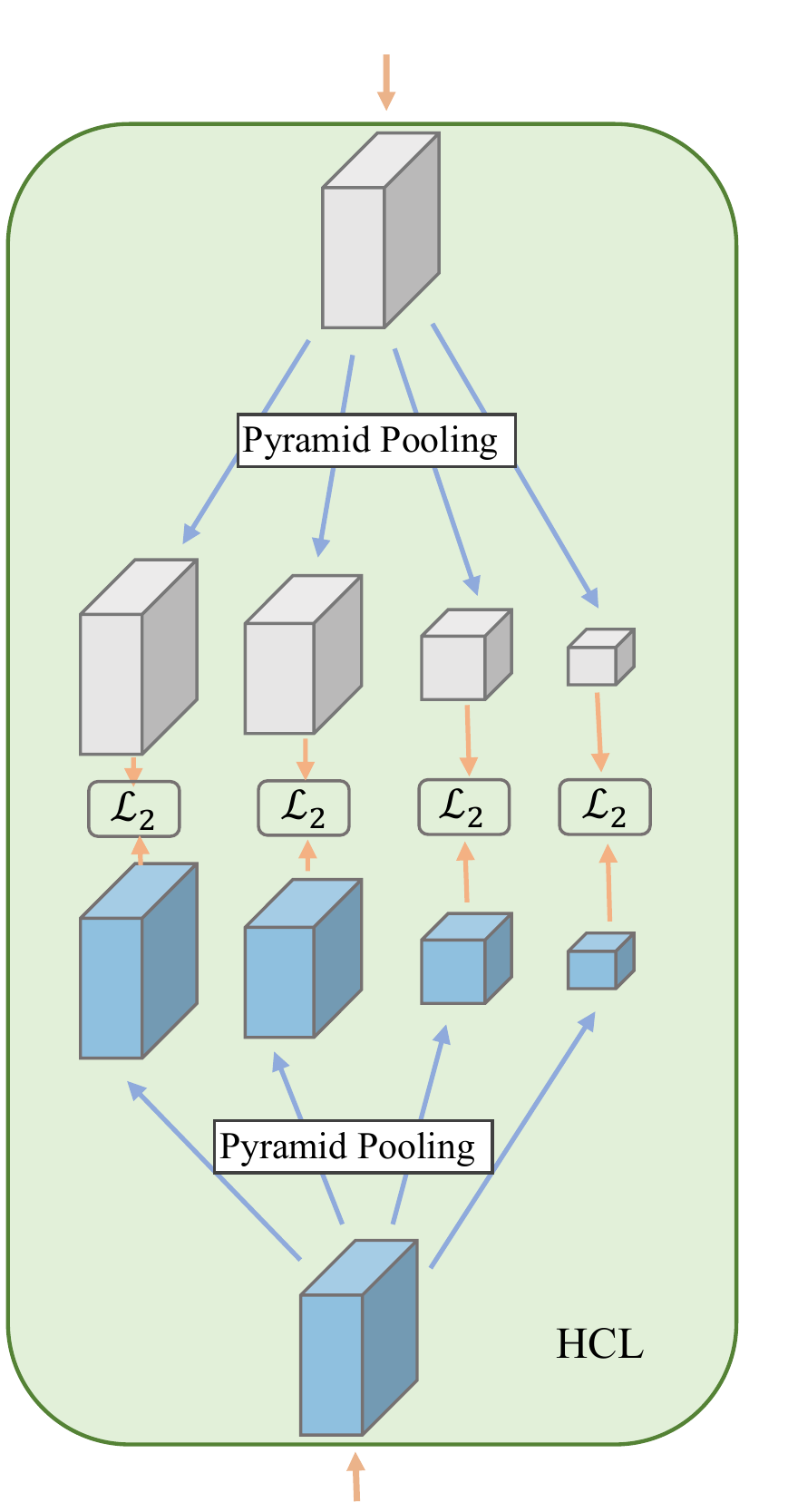}
			\label{fig:hcl}		
		\end{minipage}
	}
	\caption{(a) Architecture of ABF. Different levels' features of the student are aggregated together with attention maps. (b) Architecture of HCL. The student and teacher's features are pyramid pooled to extract different context information to distill. }
	\label{fig:detail}
	\vspace{-0.1in}
\end{figure}

\section{Our Method}

We first formalize the knowledge distillation process and the review mechanism. Then we propose a novel framework and introduce attention based fusion module and hierarchical context loss function.

\subsection{Review Mechanism}
\label{sec:rm}

Given an input image $\mathbf{X}$ and student network $\mathcal{S}$, we let $\mathbf{Y}_s=\mathcal{S}(\mathbf{X})$ represent the output logit of the student. $\mathcal{S}$ can be separated into different parts $(\mathcal{S}_1,\mathcal{S}_2,\cdots,\mathcal{S}_n,\mathcal{S}_c)$, where $\mathcal{S}_c$ is the classifier and $\mathcal{S}_1,\cdots,\mathcal{S}_n$ are different stages separated by downsample layers. Thus, the process of generating output $\mathbf{Y}_s$ can be denoted as
\begin{align}
\mathbf{Y}_s = \mathcal{S}_c \circ \mathcal{S}_n \circ \cdots \circ \mathcal{S}_1 (\mathbf{X}).
\end{align}
We refer to ``$\circ$'' as nesting of functions where $g \circ f(x) = g(f(x))$. $\mathbf{Y}_s$ is the output of student, and intermidate features are $(\mathbf{F}_s^1,\cdots,\mathbf{F}_s^n)$. The $i$th feature is calculated as
\begin{align}
\mathbf{F}_s^i = \mathcal{S}_i \circ \cdots \circ \mathcal{S}_1 (\mathbf{X}).	
\end{align}
For the teacher network $\mathcal{T}$, the process is almost the same and we omit the details. Following previous notations, single-layer knowledge distillation can be represented as 
\begin{align}
\mathcal{L}_{SKD} = \mathcal{D}\left(\mathcal{M}_s^i(\mathbf{F}_s^i), \mathcal{M}_t^i(\mathbf{F}_t^i)\right),
\end{align}
where $\mathcal{M}$ is transformation that transfers the feature to target representation of attention maps \cite{at} or factors \cite{ft}. $\mathcal{D}$ is the distance function that measures the gap between the student teacher. Similarly, multiple-layers knowledge distillation is written as
\begin{align}
\mathcal{L}_{MKD} = \sum_{i \in \mathbf{I}} \mathcal{D}\left(\mathcal{M}_s^i(\mathbf{F}_s^i), \mathcal{M}_t^i(\mathbf{F}_t^i)\right),
\end{align}
where $\mathbf{I}$ stores the layers of features to transfer knowledge.

Our {\it review} mechanism is to use previous features to guide the current feature. The single-layer knowledge distillation with the review mechanism is formalized as
\begin{align}
\mathcal{L}_{SKD\_R} = \sum_{j=1}^i \mathcal{D}\left(\mathcal{M}_s^{i,j}(\mathbf{F}_s^i), \mathcal{M}_t^{j,i}(\mathbf{F}_t^j)\right).
\end{align}
Although at the first glance it shares some similarity with multiple-layers knowledge distillation, it is in fact fundamentally different. Here feature of the student is fixed to $\mathbf{F}_s^i$, and we use the teacher's first $i$ levels of features to guide $\mathbf{F}_s^i$. The review mechanism and multiple-layers distillation are complementary concepts. When combining the review mechanism with multiple-layers knowledge distillation, the loss function becomes
\begin{align}
\mathcal{L}_{MKD\_R} = \sum_{i \in \mathbf{I}} \left( \sum_{j=1}^i \mathcal{D}\left(\mathcal{M}_s^{i,j}(\mathbf{F}_s^i), \mathcal{M}_t^{j,i}(\mathbf{F}_t^j)\right) \right).
\label{eq:mkdr}
\end{align}
In our experiments, the $\mathcal{L}_{MKD\_R}$ loss is simply added alone with original losses during the training process, and the inference is exactly the same as the original model. So our method is totally {\it cost-free at test time}. We use factor $\lambda$ to balance the distillation loss and original losses. Taking the classification task as an example, the whole loss function is defined as
\begin{align}
\mathcal{L} = \mathcal{L}_{CE} + \lambda \mathcal{L}_{MKD\_R} .
\end{align}
In our proposed review mechanism, we only use shallower features of the teacher to supervise deeper features of the student. We found that the opposite brings marginal benefit and wastes many resources instead. The intuitive explanation is that deeper and more abstracted features are too complicated for early-stage learning. More analysis is in Section \ref{sec:analysis}.

\subsection{Residual Learning Framework}
\label{sec:rlf}
Following previous work, we first design a straightforward framework, as shown in Figure \ref{fig:imp1}. 
The transformation $\mathcal{M}_s^{i,j}$ is simply composed of convolution layers and nearest interpolation layers to transfer the $i$th feature of the student to match the size of teacher's $j$th feature. We do not transform teacher features $\mathbf{F}_t$. 
The student feature is transformed into the same size as the teacher features. 

Figure \ref{fig:imp2} shows directly applying the idea to multiple-layer distillation with all-stage features distilled. However, this strategy is not optimal because of the huge information difference between stages. 
Also, it yields a complicated process where all features are used. For instance, a network with $n$ stages needs to calculate $n(n+1)/2$ pairs of features regarding the loss functions, which makes the learning process cumbersome and costs many resources.

To make the procedure more feasible and elegant, we reformulate Eq. \eqref{eq:mkdr} for Figure \ref{fig:imp2} as
\begin{align}
\mathcal{L}_{MKD\_R} = \sum_{i=1}^n \left( \sum_{j=1}^i \mathcal{D}\left(\mathbf{F}_s^i, \mathbf{F}_t^j\right) \right).
\end{align}
where the transform of features is omited for simplicity. 
We now switch the order of two summations of $i$ and $j$ as
\begin{align}
\mathcal{L}_{MKD\_R} = \sum_{j=1}^n \left( \sum_{i=j}^n \mathcal{D}\left(\mathbf{F}_s^i, \mathbf{F}_t^j\right) \right).
\label{eq:9}
\end{align}
When $j$ is fixed, Eq. \eqref{eq:9} accumulates the distance between the teacher feature $\mathbf{F}_t^j$ and student features $\mathbf{F}_s^j$-$\mathbf{F}_s^n$. With fusion of features \cite{exfuse,fpn}, we approximate the summation of distance as the distance of fused features. It leads to
\begin{align}
\sum_{i=j}^n \mathcal{D}\left(\mathbf{F}_s^i, \mathbf{F}_t^j\right) \approx \mathcal{D}\left(\mathcal{U}(\mathbf{F}_s^j,\cdots,\mathbf{F}_s^n),\mathbf{F}_t^j\right),
\end{align}
where $\mathcal{U}$ is a module to fuse features. This approximation is illustrated in Figure \ref{fig:imp3} where the structure is more effective now. But the calculation of fusion can be further optimized in a progressively manner as shown in Figure \ref{fig:imp4} for higher efficiency. Fusion of $\mathbf{F}_s^j,\cdots,\mathbf{F}_s^n$ is calculated by combination of $\mathbf{F}_s^j$ and $\mathcal{U}(\mathbf{F}_s^{j+1},\cdots,\mathbf{F}_s^n)$, where the fusion operation is recursively defined as $\mathcal{U}(\cdot, \cdot)$, applied to consecutive feature maps. Denoting $\mathbf{F}_s^{j+1,n}$ as fusion of features from $\mathbf{F}_s^{j+1}$ to $\mathbf{F}_s^{n}$, the loss is written as 
\begin{align}
\mathcal{L}_{MKD\_R} = \mathcal{D}(\mathbf{F}_s^n, \mathbf{F}_t^n)
+\sum_{j=n-1}^1 \mathcal{D}\left(\mathcal{U}(\mathbf{F}_s^j,\mathbf{F}_s^{j+1,n}),\mathbf{F}_t^j\right),
\label{eq:11}
\end{align}
Here we loop from $n-1$ down to 1 to make use of $\mathbf{F}_s^{j+1,n}$. $\mathbf{F}_s^{n,n} = \mathcal{M}_s^{n,n}(\mathbf{F}_s^n)$. The detailed structure is shown in Figure \ref{fig:imp4}, where ABF and HCL are fusion module and loss function designed for this structure, respectively. Their details are discussed in Section \ref{sec:abfhcl}.

\begin{table*}[t]
	\centering

	\begin{tabular}{c @{\hspace{0.3in}} c @{\hspace{0.25in}} c  c  c  c  c  c }
		\toprule
		\multirow{4}{*}{\makecell{Distillation \\ Mechanism}} & Teacher & ResNet56 & ResNet110 & ResNet32x4 & WRN40-2 & WRN40-2 & VGG13  \\
		& Acc     & 72.34    & 74.31     & 79.42      & 75.61   & 75.61   & 74.64  \\ 
		\cmidrule{2-8}
		& Student & ResNet20 & ResNet32  & ResNet8x4  & WRN16-2 & WRN40-1 & VGG8   \\
		& Acc     & 69.06    & 71.14     & 72.50      & 73.26   & 71.98   & 70.36  \\
		\midrule
		Logits & KD \cite{kd}      & 70.66    & 73.08     & 73.33      & 74.92   & 73.54   & 72.98  \\
		\midrule
		Single Layer & FitNet \cite{fitnet}  & 69.21    & 71.06     & 73.50      & 73.58   & 72.24   & 71.02  \\
		Single Layer & PKT \cite{pkt}     & 70.34    & 72.61     & 73.64      & 74.54   & 73.54   & 72.88  \\
		Single Layer & RKD \cite{rkd}     & 69.61    & 71.82     & 71.90      & 73.35   & 72.22   & 71.48  \\
		Single Layer & CRD \cite{crd}     & 71.16    & 73.48     & 75.51      & 75.48   & 74.14   & 73.94  \\
		\midrule
		Multiple Layers & AT \cite{at}      & 70.55    & 72.31     & 73.44      & 74.08   & 72.77   & 71.43  \\
		Multiple Layers & VID \cite{vid}      & 70.38    & 72.61     & 73.09      & 74.11   & 73.30   & 71.23  \\
		Multiple Layers & OFD \cite{ofd}     & 70.98    & 73.23     & 74.95      & 75.24   & 74.33   & 73.95  \\
		\midrule
		\midrule
		Review & Ours&\textbf{71.89}&\textbf{73.89}&\textbf{75.63}&\textbf{76.12}&\textbf{75.09}&\textbf{74.84}\\
		\bottomrule
	\end{tabular}
	\vspace{0.1in}
	\caption{Results on CIFAR-100. The teacher and student have architectures of the same style.}
	\label{tab:c100s}
\end{table*}

\begin{table*}[tb]
	\centering
	\begin{tabular}{c c  c  c  c  c  c }
		\toprule
		\multirow{4}{*}{\makecell{Distillation \\ Mechanism}} & Teacher & ResNet32x4  & WRN40-2     & VGG13       & ResNet50   & ResNet32x4   \\
		&Acc     & 79.42       & 75.61       & 74.64       & 79.34      & 79.42        \\ 
		\cmidrule{2-7}
		&Student &ShuffleNetV1 &ShuffleNetV1 &MobileNetV2  &MobileNetV2 &ShuffleNetV2  \\
		&Acc     & 70.50       & 70.50       & 64.6        & 64.6       & 71.82        \\
		\midrule
		Logits & KD \cite{kd}      & 74.07       & 74.83       & 67.37       & 67.35      & 74.45        \\
		\midrule
		Single Layer &FitNet \cite{fitnet}  & 73.59       & 73.73       & 64.14       & 63.16      & 73.54        \\
		Single Layer &PKT \cite{pkt}     & 74.10        & 73.89       & 67.13       & 66.52      & 74.69        \\
		Single Layer &RKD \cite{rkd}     & 72.28       & 72.21       & 64.52       & 64.43      & 73.21        \\
		Single Layer &CRD \cite{crd}     & 75.11       & 76.05       & 69.73       & 69.11      & 75.65        \\
		\midrule
		Multiple Layers &AT \cite{at}      & 71.73       & 73.32       & 59.40       & 58.58      & 72.73        \\
		Multiple Layers&VID \cite{vid}      & 73.38       & 73.61       & 65.56       & 67.57      & 73.40        \\
		Multiple Layers&OFD \cite{ofd}     & 75.98       & 75.85       & 69.48       & 69.04      & 76.82        \\
		\midrule
		\midrule
		Review &Ours&\textbf{77.45}&\textbf{77.14}&\textbf{70.37}&\textbf{69.89}&\textbf{77.78}\\
		\bottomrule
	\end{tabular}
	\vspace{0.1in}
	\caption{Results on CIFAR-100 with the teacher and student having different architectures.}
	\label{tab:c100d}
\end{table*}

The structure in Figure \ref{fig:imp4} is elegant and eases the distillation process with utilizing the concept of residual learning. 
For instance, the stage-4's feature of the student is aggregated with stage-3's feature of the student to mimic the stage-3's feature of the teacher. Therefore, stage-4's feature of the student learns the residual of stage-3's feature between the student and teacher. The residual information is very likely to be the key factor that the teacher yields higher-quality results.

This residual learning process is more stable and effective than directly letting high-level features of the student learned from low-level features of the teacher. 
With the residual learning framework, the high-level features of the student can better extract useful information progressively.
Further, using Eq. \eqref{eq:11}, we eliminate the summation and reduce the total complexity to $n$ pairs of distances.

\begin{table*}[t]
	\centering
	\begin{tabular} {c | c c c c c c c c}
		\toprule
		Setting & & Teacher & Student & KD \cite{kd} & AT \cite{at} & OFD \cite{ofd} & CRD \cite{crd} & Ours \\
		\midrule
		\multirow{2}*{(a)}&Top-1 & 76.16 & 68.87 & 68.58 & 69.56 & 71.25 & 71.37 & \textbf{72.56} \\
		&Top-5 & 92.86 & 88.76 & 88.98 & 89.33 & 90.34 & 90.41 & \textbf{91.00} \\
		\midrule
		\multirow{2}*{(b)}&Top-1 & 73.31 & 69.75 & 70.66 & 70.69 & 70.81 & 71.17 & \textbf{71.61} \\
		&Top-5 & 91.42 & 89.07 & 89.88 & 90.01 & 89.98 & 90.13 & \textbf{90.51} \\
		\bottomrule
	\end{tabular}
	\vspace{0.1in}
	\caption{Results on ImageNet. (a) MobileNet as student, ResNet50 as teacher. (b) ResNet18 as student, ResNet34 as teacher.}
	\label{tab:imagenet}
\end{table*}

\subsection{ABF and HCL}
\label{sec:abfhcl}

There are two key components in Figure \ref{fig:imp4}. They are attention based fusion (ABF) and hierarchical context loss (HCL). We explain them here. 

ABF module utilizes the insight of \cite{attention,senet}, as shown in Figure \ref{fig:abf}. The higher level features are first resized to the same shape as the lower level features. Then two features from different levels are concatenated together to generate two $H\times W$ attention maps. These maps are multiplied with two features, respectively. Finally, the two features are added to generate the final output.

The ABF module can generate different attention maps according to  input features. So the two feature maps can be dynamically aggregated. The adaptive sum is better than direct sum because the two feature maps are from different stages of the network and their information is diverse. The low-  and high-level features may focus on different partitions. The attention maps can aggregate them more reasonably. More experimental results are included in Section \ref{sec:abl}.

The detail of HCL is shown in Figure \ref{fig:hcl}. Usually, we use $\mathcal{L}_2$ distance as the loss function between the two feature maps. The $\mathcal{L}_2$ distance is effective to transfer information between features from the same level. But in our framework, different levels' information is aggregated together to learn from the teacher. The trivial global $\mathcal{L}_2$ distance is not powerful enough to transfer compound levels' information. 

Inspired by \cite{pspnet}, we propose HCL, utilizing spatial pyramid pooling, to separate the transfer of knowledge into different levels' context information. In this way, the information is better distilled in different abstract levels. The structure is very simple: we first extract different levels' knowledge from the feature using spatial pyramid pooling, and then use $\mathcal{L}_2$ distance to distill between them respectively. Despite the simple structure, HCL is suitable for our framework. More experimental results are shown in Section \ref{sec:abl}.

\section{Experiments}
\label{sec:experiments}
We conduct experiments on various tasks. First, we compare our method with other knowledge distillation ones regarding classification. We experiment with different settings varying architecture and datasets. Also, we apply our method to the object detection and instance segmentation tasks. Our method also improves the baseline model by large margins consistently. 

\subsection{Classification}

\paragraph{Datasets}
(1) CIFAR-100 contains 50K training images with 0.5K images per class and 10K test images.  (2) ImageNet \cite{imagenet} is the most challenging dataset for classification, which provides 1.2 million images for training and 50K images for validation over 1,000 classes.

\vspace{-0.1in} \paragraph{Implementation Details}
On CIFAR-100 dataset, we experiment with different representative network architectures, including VGG \cite{vgg}, ResNet \cite{resnet}, WideResNet \cite{wideresnet}, MobileNet \cite{mobilev2}, and ShuffleNet \cite{shufflev1,shufflev2}. We use the same training setting of \cite{crd}, except for linearly scaling up the initial learning rate and setting batch size following \cite{onehour}. 

Specifically, we train all models for 240 epochs with learning rate decayed by 0.1 for every 30 epochs after the first 150 epochs. We initialize the learning rate to 0.02 for MobileNet and ShuffleNet, and 0.1 for other models. The batch size is 128 for all models. We train all models for three times and report the mean accuracy. For fairness, previous method results are either reported in previous papers (when the training setting is the same as ours) or obtained using author released codes with our training setting.

On ImageNet, we use the standard training process that trains the model for 100 epochs and decays the learning rate for every 30 epochs. We initialize learning rate to 0.1 and set batch size to 256.

\begin{table*}
	\centering
	\begin{tabular}{l l | l @{\hspace{0.3in}} c@{\hspace{0.3in}} c@{\hspace{0.3in}} c@{\hspace{0.3in}} c@{\hspace{0.3in}} c }
		\toprule
		& Method & mAP & AP50 & AP75 & APl & APm & APs \\
		\midrule
		Teacher                &Faster R-CNN w/ R101-FPN &42.04 &62.48 &45.88 &54.60 &45.55 &25.22 \\
		Student &Faster R-CNN w/ R18-FPN  &33.26 &53.61 &35.26 &43.16 &35.68 &18.96 \\
		& ~ w/ KD \cite{kd}           &33.97 (+0.61) &54.66 &36.62 &44.14 &36.67 &18.71 \\
		& ~ w/ FitNet \cite{fitnet}   &34.13 (+0.87)&54.16 &36.71 &44.69 &36.50 &18.88 \\
		& ~ w/ FGFI \cite{finegrained} &35.44 (+2.18)&55.51 &38.17 &47.34 &38.29 &19.04\\
		& ~ w/ Our Method     &\textbf{36.75 (+3.49)} &\textbf{56.72} &\textbf{34.00} &\textbf{49.58} &\textbf{39.51} &\textbf{19.42} \\
		\midrule
		Teacher                &Faster R-CNN w/ R101-FPN &42.04 &62.48 &45.88 &54.60 &45.55 &25.22 \\
		Student &Faster R-CNN w/ R50-FPN  &37.93 &58.84 &41.05 &49.10 &41.14 &22.44 \\
		& ~ w/ KD \cite{kd}           &38.35 (+0.42) &59.41 &41.71 &49.48 &41.80 &22.73 \\
		& ~ w/ FitNet \cite{fitnet}   &38.76 (+0.83) &59.62 &41.80 &50.70 &42.20 &22.32\\
		& ~ w/ FGFI \cite{finegrained} &39.44 (+1.51) &60.27 &43.04 &51.97 &42.51 &22.89 \\
		& ~ w/ Our Method &\textbf{40.36 (+2.43)} &\textbf{60.97} &\textbf{44.08} &\textbf{52.87} &\textbf{43.81} &\textbf{23.60} \\
		\midrule
		Teacher                & Faster R-CNN w/ R50-FPN &40.22 &61.02 &43.81 &51.98 &43.53 &24.16 \\
		Student & Faster R-CNN w/ MV2-FPN &29.47 &48.87 &30.90 &38.86 &30.77 &16.33 \\
		& ~ w/ KD \cite{kd}   &30.13 (+0.66) &50.28 &31.35 &39.56 &31.91 &16.69 \\
		& ~ w/ FitNet \cite{fitnet}   &30.20 (+0.73) &49.80 &31.69 &39.69 &31.64 &16.39 \\
		& ~ w/ FGFI \cite{finegrained} &31.16 (+1.69)&50.68 &32.92 &42.12 &32.63 &16.73 \\
		& ~ w/ Our Method &\textbf{33.71 (+4.24)} &\textbf{53.15} &\textbf{36.13} &\textbf{46.47} &\textbf{35.81} &\textbf{16.77} \\
		\midrule
		Teacher                & RetinaNet101       &40.40 &60.25 &43.19 &52.18 &44.34 &24.03 \\
		Student & RetinaNet50        &36.15 &56.03 &38.73 &46.95 &40.25 &21.37 \\
		& ~ w/ KD \cite{kd} 			 &36.76 (+0.61) &56.60 &39.40 &48.17 &40.56 &21.87 \\
		& ~ w/ FitNet \cite{fitnet}   &36.30 (+0.15) &55.95 &38.95 &47.14 &40.32 &20.10 \\
		& ~ w/ FGFI \cite{finegrained}        &37.29 (+1.14) &57.13 &40.04 &49.71 &41.47 &21.01 \\
		& ~ w/ Our Method &\textbf{38.48 (+2.33)} &\textbf{58.22} &\textbf{41.46} &\textbf{51.15} &\textbf{42.72} &\textbf{22.67} \\
		\bottomrule
	\end{tabular}
	\vspace{0.1in}
	\caption{Results on object detection. We use AP on different settings to evaluate results. R101 represents using ResNet101 as backbone, and MV2 stands for MobileNetV2.}
	\label{tab:detection}
\end{table*}

\begin{table*}[t]
	\centering
	\begin{tabular}{l l | l @{\hspace{0.3in}} c@{\hspace{0.3in}} c@{\hspace{0.3in}} c@{\hspace{0.3in}} c@{\hspace{0.3in}} c }
		\toprule
		& Method & mAP & AP50 & AP75 & APl & APm & APs \\
		\midrule
		Teacher                &Mask R-CNN w/ R101-FPN &38.63 &60.45 &41.28 &55.29 &41.33 &19.48 \\
		\multirow{2}*{Student} &Mask R-CNN w/ R18-FPN  &31.25 &51.07 &33.10 &45.53 &32.80 &14.18 \\
		& ~ + Our Method &\textbf{33.62 (+2.37)} &\textbf{53.91} &\textbf{35.96} &\textbf{50.30} &\textbf{35.31} &\textbf{15.03}\\
		\midrule
		Teacher                &Mask R-CNN w/ R101-FPN &38.63 &60.45 &41.28 &55.29 &41.33 &19.48 \\
		\multirow{2}*{Student} &Mask R-CNN w/ R50-FPN  &35.24 &56.32 &37.49 &50.34 &37.71 &17.16 \\
		& ~ + Our Method &\textbf{36.98 (+1.74)} &\textbf{58.13} &\textbf{39.60} &\textbf{53.19} &\textbf{39.57} &\textbf{17.54} \\
		\midrule
		Teacher                &Mask R-CNN w/ R50-FPN  &37.17 &58.60 &39.88 &53.30 &39.49 &18.63 \\
		\multirow{2}*{Student} &Mask R-CNN w/ MV2-FPN  &28.37 &47.19 &29.95 &41.70 &29.01 &12.09 \\
		& ~ + Our Method &\textbf{31.56 (+3.19)} &\textbf{50.70} &\textbf{33.44} &\textbf{47.39} &\textbf{32.44} &\textbf{12.76} \\
		\bottomrule
	\end{tabular}
	\vspace{0.1in}
	\caption{Instance segmentation results. R101 and MV2 stand for ResNet101 and MobileNetV2.}
	\label{tab:ins}
\end{table*}

\vspace{-0.1in} \paragraph{Results on CIFAR-100} 
Table \ref{tab:c100s} summarizes results on CIFAR-100 with the teacher and student having architectures of the same style. We separate previous methods in different groups according to the features they use. KD is the only method that uses logits. Methods in FitNet group use single-layer information, and methods in AT group use multiple-layer information. Our method employs multi-layer feature with the review mechanism. It outperforms all previous methods on all architectures. 

We also experiment with the setting that the student and teacher have different architectural styles, and show results in Table \ref{tab:c100d}.
Method of OFD \cite{ofd} and ours use multiple layers for distillation. They outperform those with distillation from the last layer, manifesting that our knowledge review mechanism successfully relaxes previously emphasized intermediate- or last-layer distillation condition \cite{crd}. 

\vspace{-0.1in} \paragraph{Results on ImageNet} The number of images in CIFAR-100 is small. So we also conduct experiments on ImageNet to verify the scalability of our method. We
experiment with two settings of distillation from ResNet50 to MobileNet \cite{mobilenet}, and from ResNet34 to ResNet18  respectively. Our method, again, outperforms all other methods, as reported in Table \ref{tab:imagenet}. Setting (a) is challenging due to architecture difference. But the advantage of our method is consistently prominent. On setting (b), gap between the student and teacher is already reduced to a very small value 2.14 by previous best method. We further reduce it to 1.70, achieving 20\% relative performance improvement.

\subsection{Object Detection}
We also apply our method to other computer vision tasks. On object detection, like the procedure for the classification task, we distill between the student and teacher's backbone output features. More details are presented in the supplementary file. We use the representative COCO2017 dataset \cite{coco} to evaluate our method and take the most popular open-source report Detectron2 \cite{wu2019detectron2} as our strong baseline. We use the best pre-trained model provided by Detrctron2 as teacher. Student models are trained using the standard training policy following tradition \cite{finegrained}. All performance is evaluated on COCO2017 validation set. We conduct experiments on both two- and one-stage methods.

Since only a few methods \cite{finegrained,ofd} are claimed workable for detection, we reproduce the popular ones \cite{kd,fitnet} and the latest one \cite{finegrained}. The comparison is presented in Table \ref{tab:detection}. We note that knowledge distillation methods, such as KD and FitNet, also improve the performance of detection. But the gain is limited. FGFI \cite{finegrained} is directly designed for detection, and works better than other methods on this task. Still, our method outperforms it by a large margin. 

We also vary experimental setting to check the generality. On the two-stage method FasterRCNN \cite{faster}, we change backbone architectures. The knowledge distillation between architectures of the same style boosts mAP of ResNet18 and ResNet50 by 3.49 and 2.43 respectively. They are significant numbers. The distillation between ResNet50 and MobileNetV2 still promotes the baseline from 29.47 to 33.71. On one-stage detector RetinaNet \cite{retinanet}, the gap between student and teacher is small, our method also improves the mAP by 2.33. The success on challenging object detection tasks demonstrates the generality and effectiveness of our method.

\newcommand{\tred}{\textcolor{red}}

\newcommand{\tgreen}{\textcolor{blue}}

\begin{table}
	\centering
	\begin{tabular}{c c | c c c c}
		\toprule
		& & \multicolumn{4}{c}{Teacher Stage} \\
		& & 1 & 2 & 3 & 4 \\
		\midrule
		\multirow{4}{*}{\rotatebox{90}{Student Stage}}	
		& 1 & \tgreen{69.5} & \tred{69.0} & \tred{68.2} & \tred{66.3} \\
		\cmidrule{3-6}
		& 2 & \tgreen{69.6} & \tgreen{69.6} & \tred{61.4} & \tred{61.1}\\
		\cmidrule{3-6}
		& 3 & \tgreen{69.2} & \tgreen{69.8} & \tgreen{71.0} &\tred{50.4} \\
		\cmidrule{3-6}
		& 4 & \tgreen{69.2} & \tgreen{69.3} & \tgreen{70.3} & \tgreen{70.3} \\
		\bottomrule
	\end{tabular}
	\vspace{0.1in}
	\caption{Results of knowledge distillation between different stages of the teacher and student. The student's baseline result is 69.1. We use red color to mark numbers lower than baseline and blue for those higher than baseline. It is clear that using the lower level information of the teacher to supervise the deeper stage of the student is helpful. }
	\label{tab:analysis1}
\end{table}

\subsection{Instance Segmentation}

In this section, we apply our method to the even more challenging instance segmentation task. As far as we know, this is the first time for the knowledge distillation methods to apply to instance segmentation. We also use the strong baseline provided by Detectron2 \cite{wu2019detectron2}. We take Mask R-CNN \cite{mask} as our models and distill between different backbone architectures. The models are trained on the COCO2017 training set and are evaluated on the validation set. The results are shown in Table \ref{tab:ins}.

Our method also improves the performance of instance segmentation tasks notably. For distillation between architectures of the same style, we boost the performance of ResNet18 and ResNet50 by 2.37 and 1.74, and reduce the gap between the teacher and student by 32\% and 51\% relatively. Even for the distillation on architectures of different styles, we better MobileNetV2 by 3.19. 

The fact that our method performs decently on all image classification, object detection, and instance segmentation tasks and accomplishes all SOTA results, manifest the remarkable efficacy and applicability of our method.

\subsection{More Analysis}

\paragraph{Knowledge Distillation across Stages}
\label{sec:analysis}
We analyze the effectiveness of knowledge transfer across stages. 
We use ResNet20 as the student and ResNet56 as the teacher on CIFAR-100 dataset. There are four stages in ResNet20 and ResNet56. We choose the different stages in the student and vary stages in the teacher to supervise them. The results are summarized in Table \ref{tab:analysis1} .

These results conclude that distilling student with the same stage information from the teacher is the best solution. This is in accordance with our intuition. Further, it is intriguing to observe that information from lower layers is also helpful. But distilling from teacher's higher levels adversely affects training of the student. 

It indicates that deeper stages of the student are capable of learning useful information from lower stages of the teacher. In the other way around, deeper and more abstracted features from teacher are too complicated for early-stage of the student. This is consistent with our understanding and our proposed review mechanism, which uses shallow stages of the teacher to supervise deeper stages of the student.

\begin{table}
	\centering
	\begin{tabular}{l @{\hspace{0.2in}} l @{\hspace{0.2in}} l @{\hspace{0.2in}} l @{\hspace{0.2in}} c}
		\toprule
		RM 			& RLF 		& ABF 		& HCL 		& Accuracy (Variance)\\
		\midrule
		&			&			&			& 74.3 (5e-2)\\
		\ding{52}	&			&			&			& 75.2 (6e-2) \\
		\ding{52}	& \ding{52}	&			&			& 75.6 (6e-2) \\
		\ding{52}	& \ding{52}	& \ding{52}	&			& 76.0 (6e-2) \\
		\ding{52}	& \ding{52}	&			& \ding{52}	& 75.8 (5e-2) \\
		\ding{52}	& \ding{52}	& \ding{52}	& \ding{52}	& 76.2 (4e-2) \\
		\bottomrule
	\end{tabular}
	\vspace{0.1in}
	\caption{RM: The proposed review mechanism (Section \ref{sec:rm}). RLF: Residual learning frame work (Section \ref{sec:rlf}). ABF: Attentation based fusion module (Section \ref{sec:abfhcl}). HCL: Hierarchical context loss function (Section \ref{sec:abfhcl}).}
	\label{tab:abl}
\end{table}

\vspace{-0.1in} \paragraph{Ablation Study}
\label{sec:abl}
Ablation experiments are conducted, in which the ablation components are added one-by-one to measure their effect. The results are summarized in Table \ref{tab:abl} with accuracy and variance. We use WRN16-2 as the student and WRN40-2 as the teacher on CIFAR100 dataset. The baseline is trained with $\mathcal{L}_2$ distance between the same level's features of the student and the teacher.

With our proposed review mechanism, the result is improved over the baseline, as shown in the second line, which uses the trival structure as shown in Figure \ref{fig:imp2}. 
When we further refine the structure with the residual learning framework, the student yields larger gains. 
The attention based fusion module and hierarchical context loss function also provide great improvement when utilized separately.
And when we aggregate them together, the best results are obtained. It is surprising that they are even better than the teacher.

\section{Conclusion}
In this paper, we have studied knowledge distillation from a new perspective and accordingly proposed the review mechanism, which uses multiple layers in the teacher to supervise one layer in the student. Our method achieves significant improvement consistently on all classification, object detection and instance segmentation tasks, compared with all previous SOTA. We only use output of stages, and already accomplish decent results in general. 

For future work, we will also employ features inside a stage. Also, other loss functions will be investigated in our framework.

{\small
\bibliographystyle{ieee_fullname}
\bibliography{egbib}
}

\end{document}